\documentclass[sigconf,screen]{acmart}
\microtypesetup{expansion=false}
\setcopyright{none}
\renewcommand\footnotetextcopyrightpermission[1]{}

\usepackage{booktabs}
\usepackage{multirow}
\usepackage{tabularx}
\usepackage{enumitem}
\usepackage{array}
\usepackage{graphicx}
\usepackage{amsmath}

\newcommand{\datasetname}{TransBiolab}

\newcolumntype{Y}{>{\centering\arraybackslash}X}
\newcommand{\narrowtabwidth}{\columnwidth}

\title{TransBiolab: A Real-World Multi-View Dataset of Cluttered Transparent Biomedical Objects}

\author{Ke Ma}
\affiliation{%
  \institution{School of Artificial Intelligence and Automation, Huazhong University of Science and Technology}
  \city{Wuhan}
  \country{China}
}
\affiliation{%
  \institution{College of Design and Innovation, Tongji University}
  \city{Shanghai}
  \country{China}
}
\email{make@hust.edu.cn}
\orcid{0000-0002-2085-5028}

\author{Yifei Wang}
\affiliation{%
  \institution{School of Artificial Intelligence and Automation, Huazhong University of Science and Technology}
  \city{Wuhan}
  \country{China}
}
\email{m202573761@hust.edu.cn}
\orcid{0009-0001-5800-2098}

\author{Meng Wang}
\affiliation{%
  \institution{College of Design and Innovation, Tongji University}
  \city{Shanghai}
  \country{China}
}
\affiliation{%
  \institution{Shanghai Institute for Intelligent Autonomous Systems, Tongji University}
  \city{Shanghai}
  \country{China}
}
\email{mengwangtj@tongji.edu.cn}
\orcid{0000-0002-2293-1709}

\author{Tian Xia}
\authornote{Corresponding author.}
\affiliation{%
  \institution{School of Software and Engineering, Huazhong University of Science and Technology}
  \city{Wuhan}
  \country{China}
}
\email{tianxia@hust.edu.cn}
\orcid{0000-0003-1721-6126}

\renewcommand{\shortauthors}{Ma et al.}

\acmConference[MM '26]{Proceedings of the 34th ACM International Conference on Multimedia}{November 10--14, 2026}{Rio de Janeiro, Brazil}
\acmBooktitle{Proceedings of the 34th ACM International Conference on Multimedia (MM '26), November 10--14, 2026, Rio de Janeiro, Brazil}

\ccsdesc[500]{Computing methodologies~Vision for robotics}
\ccsdesc[500]{Computing methodologies~Reconstruction}
\ccsdesc[300]{Computing methodologies~3D imaging}

\begin{document}

\begin{abstract}
Autonomous biomedical laboratories increasingly rely on visual perception to recognize, localize, and manipulate transparent plasticware, yet high-quality real-world datasets for this setting remain limited. The scarcity of domain-relevant data is particularly restrictive in cluttered multi-object scenes, where mutual occlusion and view-dependent appearance changes remain challenging even for contemporary visual foundation models. Existing transparent-object datasets have advanced segmentation, depth, and pose estimation, but they usually do not evaluate the combined setting of multi-object clutter, occlusion, and calibrated multi-view capture that characterizes real laboratory manipulation scenes. To address this gap, we present \datasetname{}, a real-world RGB-D dataset of cluttered transparent biomedical objects captured as calibrated multi-view sequences. \datasetname{} contains 161{,}315 frames from 98 scenes and 1.03M instance annotations over 15 laboratory object types, including 6D poses, full and visible masks, depth, and per-frame camera calibration. The dataset is organized along three axes that reflect operational difficulty: object category, the total number of objects in a frame, and camera viewpoint. We further define dataset-centric benchmarks for segmentation, depth estimation and completion, and 6D pose estimation, and report a system-level robot manipulation evaluation enabled by the released annotations and calibrations. By focusing on repeated transparent instances, clutter, and multi-view laboratory capture, \datasetname{} provides a resource for segmentation, depth estimation, 6D pose estimation, and multi-view reasoning in autonomous laboratory manipulation. \href{https://dualtransparency.github.io/TransBiolab/}{Project page}.
\end{abstract}

\keywords{dataset, transparent objects, biomedical laboratory, multi-view perception, segmentation, depth estimation, 6D pose estimation}

\begin{teaserfigure}
  \centering
  \fontfamily{ptm}\selectfont 
  \includegraphics[width=\textwidth]{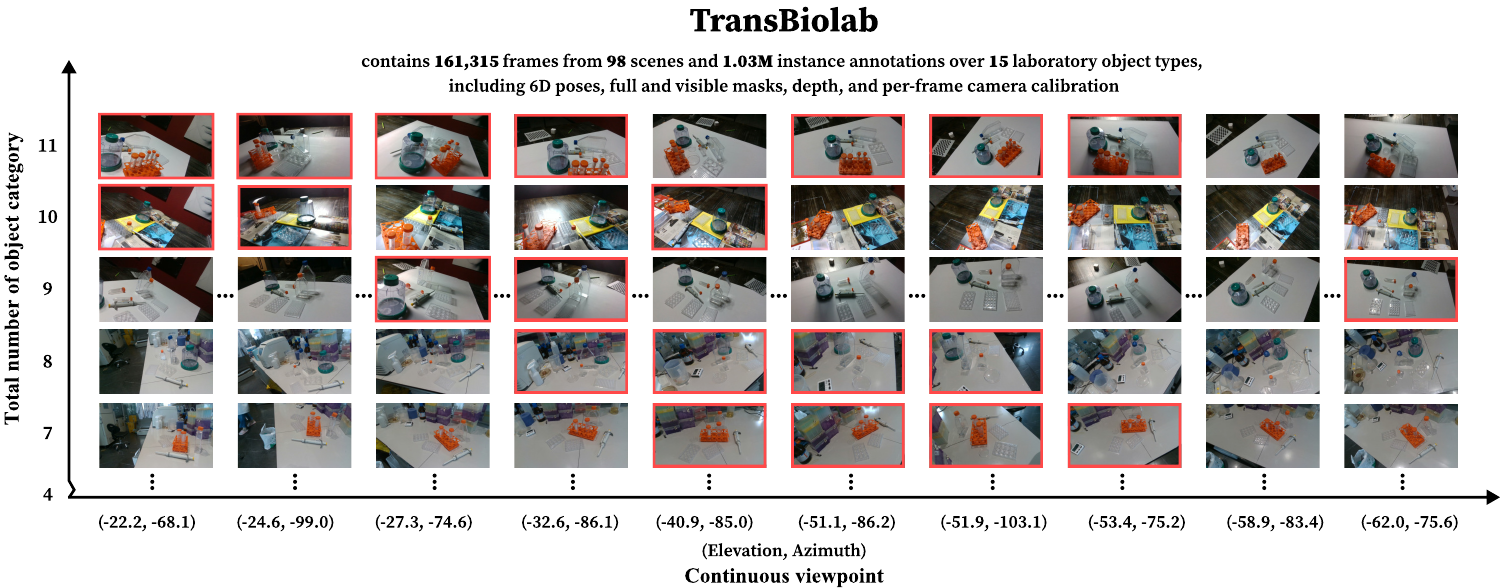}
  \caption{An overview of the proposed TransBiolab dataset, featuring multi-object and multi-view characteristics. Frames denoted by red rectangles indicate higher levels of clutter and occlusion.}
  \Description{An overview of the proposed dataset: TransBiolab.}
  \label{fig:teaser}
\end{teaserfigure}

\maketitle

\section{Introduction}

Autonomous biomedical laboratories are moving from fixed-workcell automation toward perception-driven manipulation, where robots must identify, localize, and transfer laboratory vessels across liquid handling, sample preparation, and cell-culture workflows \cite{abolhasani2023rise,tom2024selfdriving,volk2024performance,fushimi2025autonomous,melocchi2024roboticcluster,ochs2021stemcelldiscovery,moutsatsou2019automation,doulgkeroglou2020automation,zhang2023manualpipette,tan2024edtalk,tan2025fixtalk,tan2026synmotion,tan2026codance,ma2019design,shi2017digital,ma2026phys,ma2026wpis}. Many manipulated objects are transparent or semi-transparent plastic consumables, such as well plates, Petri dishes, centrifuge tubes, culture flasks, pipettes, reagent bottles, and bioreactors. Their appearance is weakly textured, strongly view dependent, and often entangled with backgrounds through reflection, refraction, and transmission \cite{sajjan2020cleargrasp,xu2022seeingglass,chen2022clearpose}.

The difficulty increases when several similar transparent objects appear together. In realistic laboratory scenes, repeated instances of plates, tubes, and dishes frequently coexist in the same frame, partially occlude each other, and interact with opaque auxiliary tools such as pipettes or racks. These effects complicate multiple perception tasks at once: segmentation must separate heavily overlapping instances, depth estimation must remain stable under missing or distorted depth returns, and 6D pose estimation must distinguish between geometrically similar objects with partial visibility \cite{liu2020keypose,liu2021stereobj,fang2022transcg}. Without data that exposes these factors jointly, it is hard to determine whether model failure comes from category ambiguity, clutter level, or viewpoint.

At the same time, perception models have become more general. FoundationPose targets unified 6D pose estimation and tracking of novel objects from object models \cite{wen2024foundationpose}; MegaPose scales object-centric pose prediction to broad object collections \cite{labbe2022megapose}; SAM~2 and SAM~3 extend promptable segmentation to images, videos, and concept-level prompts \cite{ravi2024sam2,carion2025sam3}; and recent geometry models such as Depth Anything V2 and Depth Anything~3 broaden generic depth and any-view scene understanding \cite{yang2024depthanythingv2,lin2025depthanything3,ma2026wpis}. Transparent objects remain a practical boundary case for these models because the optical and geometric ambiguities of transparent plasticware are weakly represented in standard benchmarks \cite{chen2022clearpose,lukezic2024transparenttracking}.

Existing datasets only partially cover this setting. Classical real-world 3D datasets such as LINEMOD, YCB, YCB-Video, T-LESS, ITODD, RU-APC, HOPE, and ROBI are central to object pose research, but they mainly emphasize opaque household objects or industrial parts \cite{hinterstoisser2012linemod,calli2015ycb,xiang2018posecnn,hodan2017tless,drost2017itodd,rennie2016ruapc,tyree2022hope,yang2021robi}. Transparent-object datasets such as KeyPose, TODD, StereOBJ-1M, ClearPose, TransCG, and Trans10K-v2 have substantially expanded transparent-object evaluation, yet they typically focus on household, chemical, or generic grasping objects rather than biomedical plasticware \cite{liu2020keypose,xu2022seeingglass,liu2021stereobj,chen2022clearpose,fang2022transcg,xie2021trans2seg}. Moreover, real laboratory manipulation scenes are not only transparent; they also combine repeated instances, cluttered layouts, and calibrated multi-view capture in ways that are tied to laboratory procedures.

Our key idea is that cluttered transparent-object perception should be evaluated along axes that reflect operational difficulty: (i) object category, (ii) the total number of objects visible in a frame, and (iii) camera viewpoint distribution. These factors directly guide data collection and annotation.

We introduce \datasetname{}, a real-world RGB-D dataset of 15 biomedical laboratory objects captured as calibrated multi-view sequences illustrated in Figure~\ref{fig:teaser}. In total, it provides 161{,}315 frames and 1.03M instance annotations, including 6D poses, masks, depth, and per-frame camera calibration. The dataset covers 88 controlled collection scenes with 11 arrangement patterns, 4 backgrounds, and 2 lighting settings, together with 10 held-out real-world laboratory scenes containing new layouts, more distractors, and different environmental conditions. Because each scene is captured as an ordered multi-view sequence, the dataset supports both frame-based evaluation and multi-view or video-based methods for resolving occlusion regarding transparent objects.


\section{Related Work}

\subsection{Real-world 3D datasets}

Real-world 3D datasets have shaped object pose estimation and robotic perception. LINEMOD and LM-O established early evaluation settings for textureless objects and occlusion \cite{hinterstoisser2012linemod,hodan2018bop}. T-LESS, ITODD, RU-APC, and IC-BIN extended benchmarking toward industrial, warehouse, and multi-instance scenes with symmetry, low texture, and clutter \cite{hodan2017tless,drost2017itodd,rennie2016ruapc,doumanoglou2016recovering}. YCB, YCB-Video, GraspNet, HOPE, and ROBI broadened object scale, scene realism, multi-view reasoning, grasp annotations, and reflective materials \cite{calli2015ycb,xiang2018posecnn,fang2020graspnet,tyree2022hope,yang2021robi}. These datasets remain essential references, but most center on opaque household objects or industrial parts rather than transparent biomedical consumables. The appearance and operational context of laboratory plasticware remain underrepresented in 3D benchmarks.

\subsection{Transparent object datasets}

Transparent-object datasets have expanded the scope of perception benchmarking beyond opaque rigid objects. KeyPose introduced multi-view annotation and stereo pose estimation for desktop transparent objects \cite{liu2020keypose}. StereOBJ-1M scaled stereo capture to mixed tabletop objects and object-pose supervision \cite{liu2021stereobj}. ClearPose provided a large-scale benchmark centered on transparent-object segmentation and pose analysis across household and chemical objects \cite{chen2022clearpose}. TransCG focused on real-world transparent-object depth completion and grasping in cluttered daily-life scenes \cite{fang2022transcg}, while Trans10K-v2 emphasized segmentation in the wild \cite{xie2021trans2seg}. These datasets have been important for transparent-object perception, yet many predominantly feature household or chemical glassware rather than plastic biomedical consumables, and they do not center evaluation on the conjunction of repeated transparent instances, clutter, and calibrated multi-view capture in real laboratory scenes.

\subsection{Foundation models for transparent-object perception}

Recent foundation models broaden perception across segmentation, geometry, and pose estimation tasks, but transparent objects remain difficult for all three. Promptable segmentation models such as SAM~2 and SAM~3 aim to generalize across open-vocabulary objects and video settings \cite{ravi2024sam2,carion2025sam3}. Generic geometry models such as Depth Anything V2 and Depth Anything~3 expand monocular and any-view scene understanding \cite{yang2024depthanythingv2,lin2025depthanything3}. FoundationPose and MegaPose extend 6D pose estimation to novel objects through object-model-driven inference and render-and-compare refinement \cite{wen2024foundationpose,labbe2022megapose}. Our motivation is not to propose another model, but to provide data that can test how general-purpose models behave on transparent laboratory objects across segmentation, depth-oriented reasoning, and 6D pose estimation, and to expose the gaps on this boundary case.

\section{The TransBiolab Dataset}

\subsection{Objects and capture setup}

\begin{figure}[t]
    \centering
    \includegraphics[width=\linewidth]{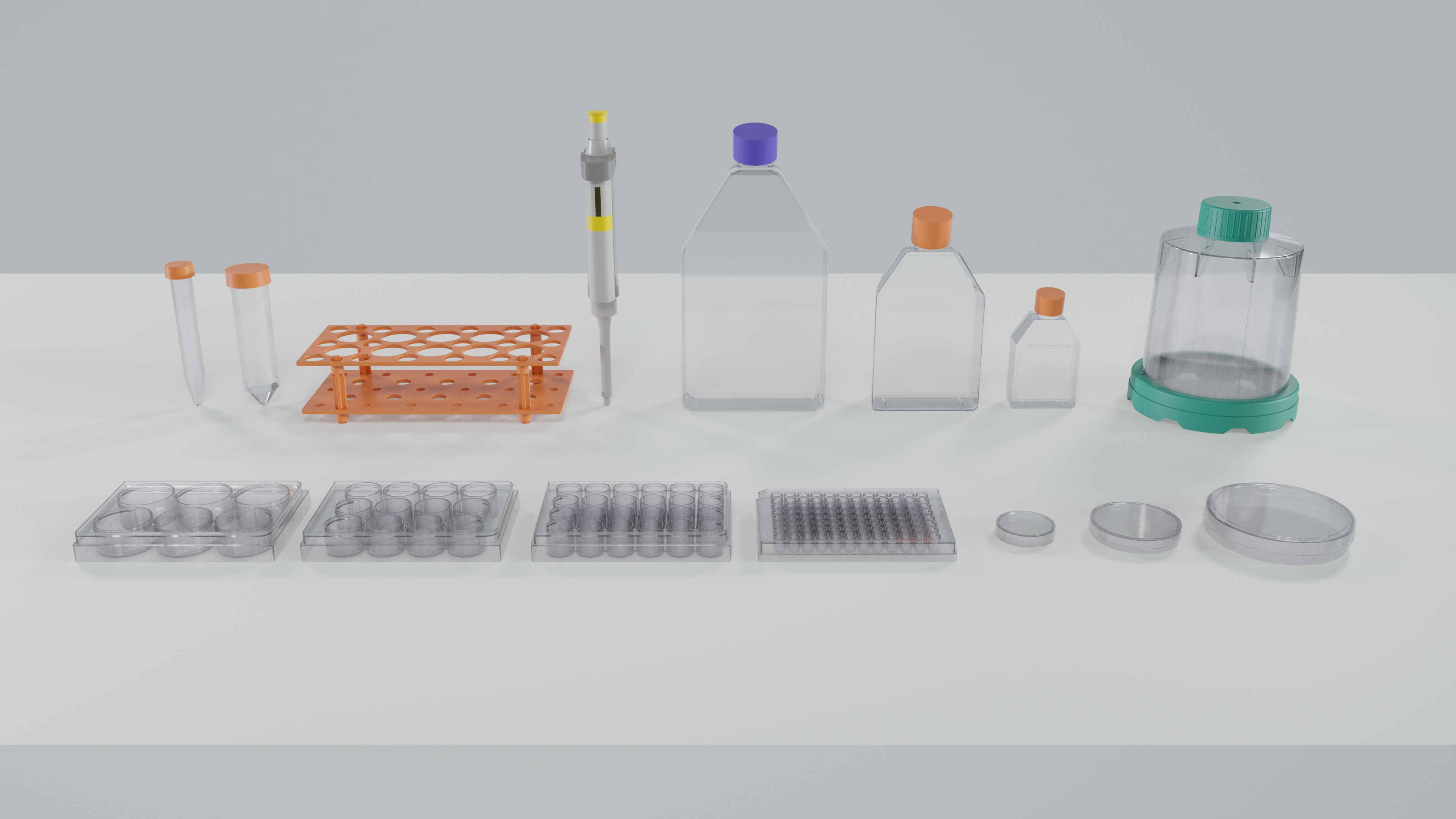}
    \caption{Rendering views of the 15 objects included in \datasetname{}, covering well plates, Petri dishes, centrifuge tubes, pipettes, tube racks, culture flasks, reagent bottles, and a bioreactor.}
    \Description{A rendered overview of 15 laboratory objects, including well plates, dishes, tubes, a rack, a pipette, flasks, bottles, and a bioreactor.}
    \label{fig:objects}
\end{figure}

\datasetname{} contains 15 laboratory objects organized into five functional groups that frequently appear in biomedical manipulation workflows: well plates (6-, 12-, 24-, and 96-well), cell culture dishes (35 mm, 60 mm, and 90 mm), liquid-handling objects (15 ml and 50 ml centrifuge tubes, pipette, and tube rack), cell culture flasks (25 ml, 75 ml, and 125 ml), and a 1 L bioreactor. Figure~\ref{fig:objects} shows the object set. Seven objects are fully transparent, six combine transparent bodies with opaque caps or bases, and two are opaque auxiliary tools. This composition reflects laboratory workcells with transparent vessels and supporting tools.

\begin{figure}[t]
    \centering
    \includegraphics[width=\linewidth]{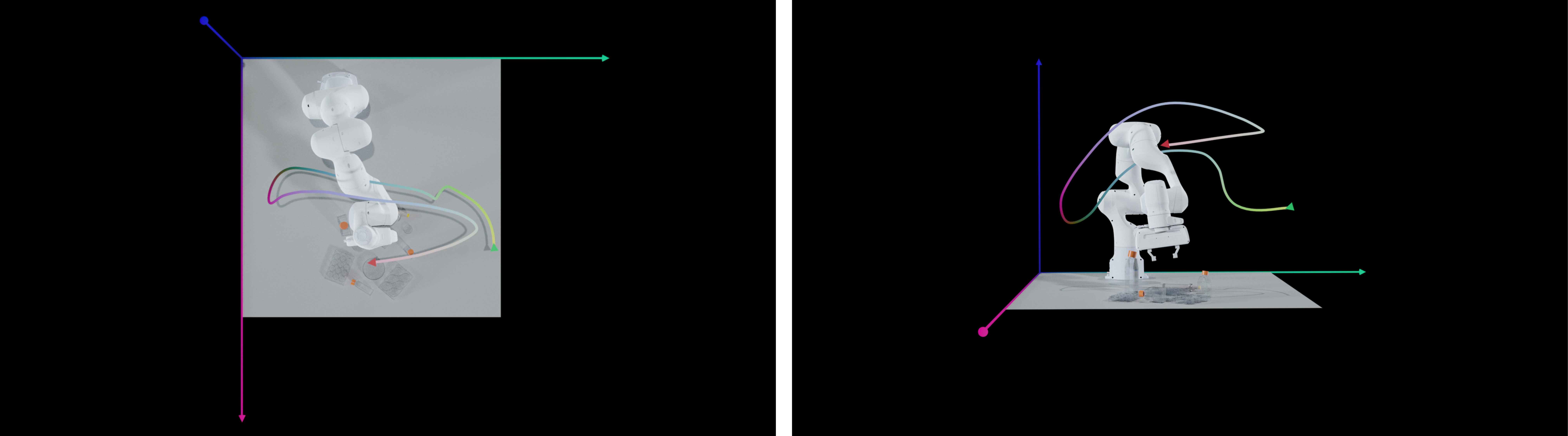}
    \caption{Calibrated robot-mounted camera trajectory from overhead and side views.}
    \Description{Two rendered views of a robot-mounted camera trajectory above a tabletop scene, shown from top and side perspectives.}
    \label{fig:trajectory}
\end{figure}

Data are collected with an Intel RealSense D435i RGB-D camera at 1280$\times$720 resolution and 30 FPS, mounted on a 7-DoF Franka Emika Panda robot. The robot-mounted setup improves repeatability and allows the camera to traverse a continuous multi-view path above the scene, as shown in Figure~\ref{fig:trajectory}. Each sequence lasts about 50 seconds and typically contains 1{,}300--1{,}800 frames; this variation follows capture timing around the nominal 30 FPS rate. The mean sequence length is about 1{,}646 frames, and the full trajectories are released as calibrated videos.

\subsection{Scene design and data acquisition}

\begin{figure}[t]
    \centering
    \includegraphics[width=\linewidth]{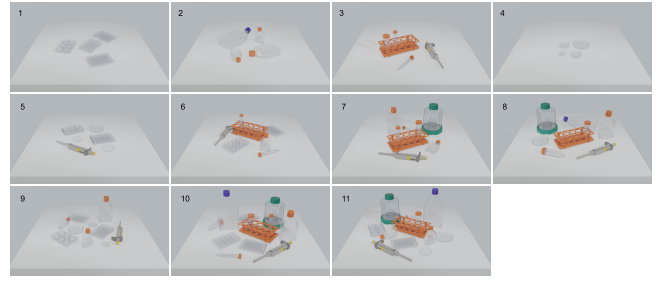}
    \caption{Eleven arrangement patterns used to build controlled scenes with increasing clutter, repeated instances, and cross-category interaction.}
    \Description{A grid showing 11 tabletop arrangement patterns with different combinations of plates, tubes, dishes, racks, bottles, and pipettes.}
    \label{fig:patterns}
\end{figure}

The 11 arrangement patterns in Figure~\ref{fig:patterns} are not random compositions. They are designed to reflect recurring laboratory procedures such as sample preparation, liquid handling, mixing, sampling, liquid transfer, transport between workstations, and cell expansion or harvest steps that require transferring materials across multiple vessels. The patterns therefore encode object combinations according to laboratory affordances and SOP-like task structure. Some patterns emphasize repeated instances from the same category, such as multiple tubes or plates, because identical vessels often appear together in practice. Other patterns mix several functional groups and explicitly include opaque auxiliary tools, such as tube racks and pipettes, because real laboratory scenes usually combine transparent and opaque objects. Across the patterns, scene difficulty increases from moderate clutter with 4--5 objects to denser scenes with 9--12 objects and heavier mutual occlusion.

\begin{figure}[t]
    \centering
    \includegraphics[width=\linewidth]{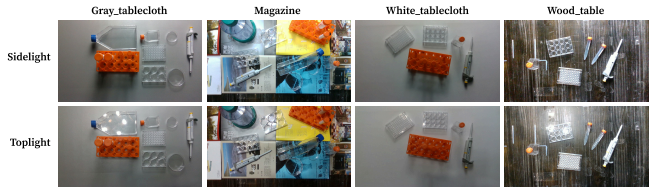}
    \caption{Controlled variations across four tabletop backgrounds and two lighting settings.}
    \Description{Examples of the same scene captured over four backgrounds and two lighting settings, plus side-view examples.}
    \label{fig:bglight}
\end{figure}

Controlled scenes are further varied by four tabletop backgrounds and two lighting settings, illustrated in Figure~\ref{fig:bglight}. The four backgrounds are a white tablecloth, a grey tablecloth, a bare wooden tabletop, and a magazine-covered tabletop. The two lighting settings are a top light of about 190 lx and a side light of about 75 lx. These factors change contrast, reflections, and occlusion cues without altering the underlying scene composition, helping isolate visual sensitivity.

Beyond the controlled tabletop capture, we collect 10 held-out real-world laboratory scenes using handheld RGB-D capture. These scenes contain new scene layouts, more distractors, and environmental changes in backgrounds and lighting, as illustrated in Figure~\ref{fig:teaser}. Across controlled and held-out scenes together, \datasetname{} contains 98 scenes, 161{,}315 RGB-D frames, and about 1.03M object instance annotations. Each scene setup contains more than three objects; 161{,}059 frames (99.84\%) contain more than three in-frame objects, while 256 frames (0.16\%) contain exactly three when camera motion briefly moves other objects out of view. We retain these frames for sequence continuity and provide an initial scene-disjoint split separating controlled scenes from the held-out laboratory scenes.


\subsection{Annotation pipeline}

Transparent objects are difficult to annotate reliably with single-frame depth alone, so we use a sequence-centric multi-view annotation pipeline. First, each object is measured at millimeter precision and reconstructed as a 3D mesh in Rhino, and we release all 15 object models in OBJ format. Next, image sequences and object models are imported into the Blender-based ProgressLabeller workspace \cite{chen2022progresslabeller}. Camera trajectories are estimated with ORB-SLAM3 \cite{campos2021orbslam3}, while the depth stream is fused into a scene-level point cloud using KinectFusion-style reconstruction. Annotators then align the object meshes in the multi-view workspace by jointly checking RGB reprojection, depth-generated point clouds, and plane consistency across views. This pipeline produces object identity, 6D pose, full mask, visible mask, and aligned depth for each frame, together with camera intrinsics and per-frame extrinsics. Sequence-level annotation takes about 40 minutes for a sequence of roughly 2{,}000 images. We additionally provide format-conversion utilities for BOP-style evaluation \cite{hodan2018bop}, so that existing pose-estimation toolchains can be adapted with limited effort.

We further audit the mutual consistency of the released CAD models, poses, masks, intrinsics, and trajectories. Projected CAD silhouettes and visible masks have a mean IoU of 0.949 (median 0.971). The median symmetric contour error is 0.62 pixels for same-view reprojection and 0.91 pixels after cross-view transfer; the 95th-percentile error is below one pixel in both cases. This geometry-consistency audit does not constitute independent metrological validation.


\subsection{Dataset visualization}

\begin{figure*}[t]
    \centering
    \includegraphics[width=\textwidth]{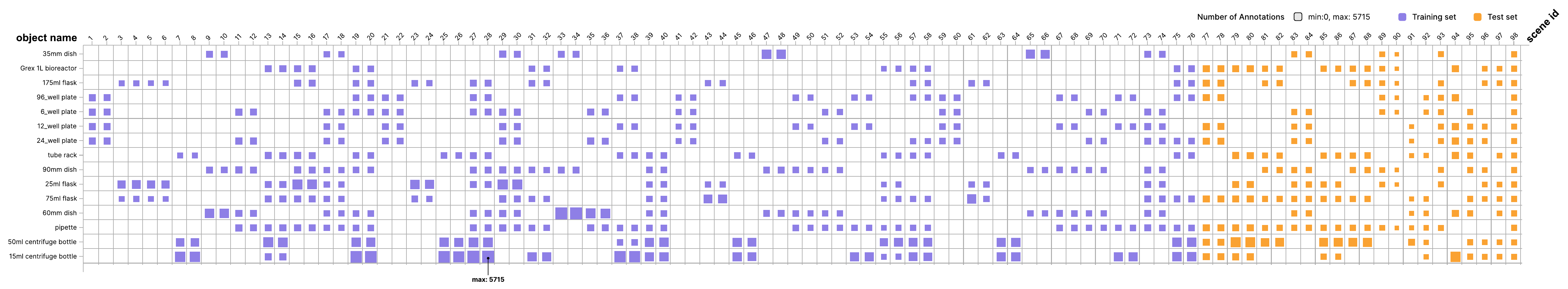}
    \caption{Scene-object distribution matrix and initial scene-disjoint training/test partition.}
    \Description{A matrix visualization showing object occurrence across scenes with a marked initial train-test partition.}
    \label{fig:matrix}
\end{figure*}

\begin{figure*}[t]
    \centering
    \includegraphics[width=0.9\textwidth]{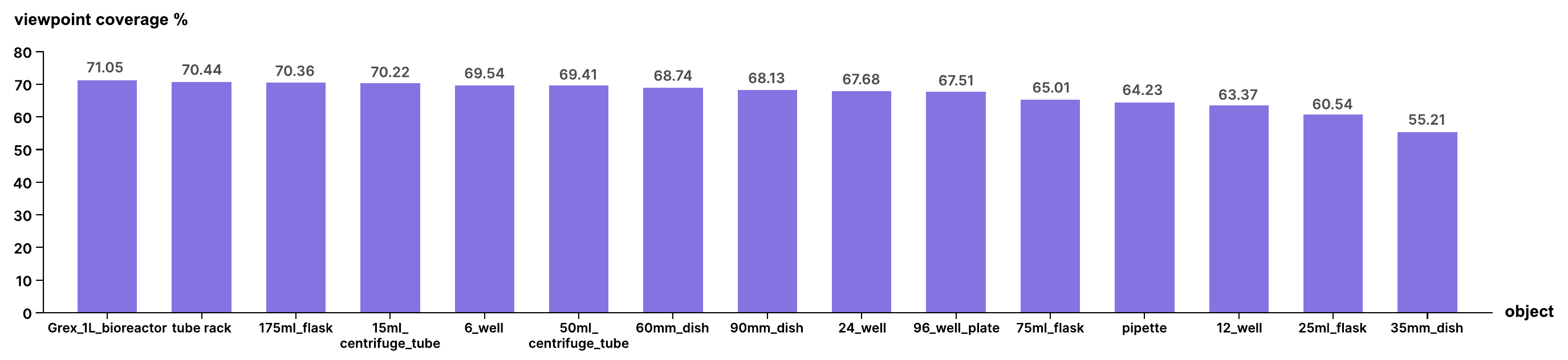}
    \caption{Object-wise viewpoint coverage across the calibrated multi-view sequences.}
    \Description{A bar chart showing viewpoint coverage for each of the 15 laboratory objects.}
    \label{fig:metadata}
\end{figure*}

\begin{figure*}[t]
    \centering
    \includegraphics[width=0.85\textwidth]{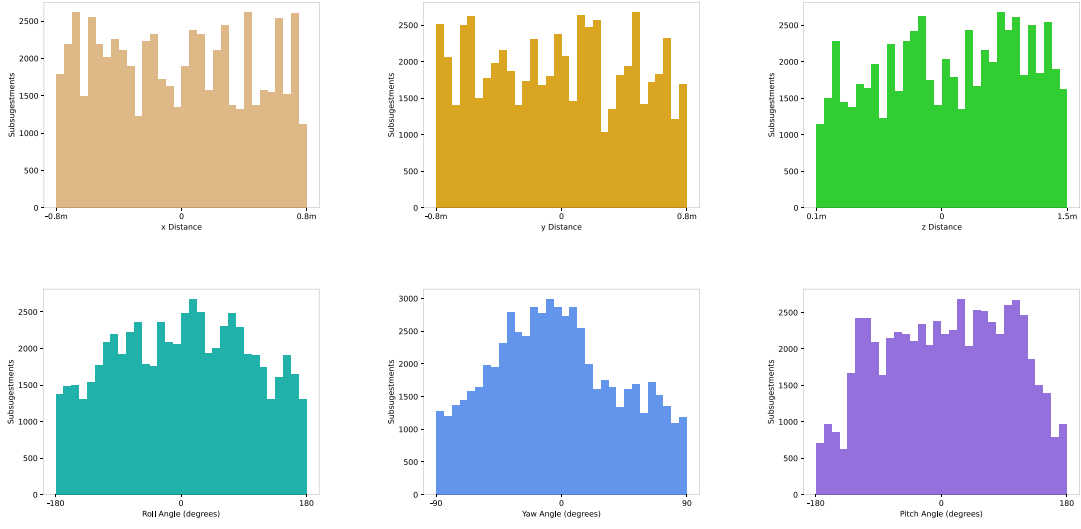}
    \caption{Distributions of released 6D annotations over translation and rotation axes.}
    \Description{Six histograms showing object translations and rotations along three axes.}
    \label{fig:6dpose}
\end{figure*}

We visualize and analyze \datasetname{} from four complementary perspectives: cross-dataset comparison, scene-object composition, calibrated multi-view structure, and released annotation distributions. Table~\ref{tab:comparison} compares representative transparent-object datasets and shows that \datasetname{} combines a biomedical laboratory focus with real-world RGB-D capture, repeated transparent instances, explicit distractors, and calibrated multi-view sequences. Figure~\ref{fig:matrix} shows the scene-object distribution and the initial train/test partition, highlighting that the 15 object types are repeatedly covered across scenes rather than concentrated in a few isolated captures. Figure~\ref{fig:teaser} illustrates the intended presentation of ten evenly sampled viewpoints from one calibrated sequence. Figure~\ref{fig:metadata} summarizes viewpoint coverage and Figure~\ref{fig:6dpose} the 6D annotation distributions across translation and rotation space, showing how object identity, camera motion, and pose labels are distributed across the dataset.

\begin{table*}[t]
\caption{Comparison with representative transparent-object datasets. N/A denotes information not explicitly reported in the primary paper.}
\label{tab:comparison}
\centering
\footnotesize
\renewcommand{\arraystretch}{1}
\setlength{\tabcolsep}{3pt}
\begin{tabularx}{\textwidth}{@{}>{\raggedright\arraybackslash}p{1.65cm}>{\raggedright\arraybackslash}p{2.30cm}>{\raggedright\arraybackslash}p{1.85cm}*{8}{>{\centering\arraybackslash}X}@{}}
\toprule
\textbf{Dataset} & \textbf{Primary domain} & \textbf{Sensor / modality} & \shortstack{\textbf{\#}\\\textbf{Objects}} & \shortstack{\textbf{\#}\\\textbf{Scenes}} & \shortstack{\textbf{\#}\\\textbf{Images}} & \shortstack{\textbf{Back-}\\\textbf{grounds}} & \textbf{Lighting} & \shortstack{\textbf{Multi-}\\\textbf{instance}} & \shortstack{\textbf{Multi-}\\\textbf{view}} & \shortstack{\textbf{Dis-}\\\textbf{tractors}} \\
\midrule
KeyPose~\cite{liu2020keypose} & Desktop transparent objects & Stereo RGB + RGB-D & 15 & 600 seq. & 48K & 10 & 1 & No & Yes & No \\
StereOBJ-1M~\cite{liu2021stereobj} & Mixed tabletop objects & Stereo RGB & 18 & 183 & 396,509 & 11 env. & Varied & Yes & Yes & Yes \\
ClearPose~\cite{chen2022clearpose} & Household + chemical & RGB-D & 63 & 51 & 354,481 & 7 sets & Varied & Yes & Yes & Yes \\
TransCG~\cite{fang2022transcg} & Daily-life transparent objects & RGB-D & 51 & 130 & 57,715 & 4 & 1 & Yes & Yes & Yes \\
Trans10K-v2~\cite{xie2021trans2seg} & Household / wild segmentation & RGB & 11 & N/A & 10,428 & Wild & Wild & Yes & No & Yes \\
\datasetname{} & Biomedical laboratory & RGB-D & 15 & 98 & 161,315 & 4 & 2 & Yes & Yes & Yes \\
\bottomrule
\end{tabularx}
\end{table*}

\section{Benchmark Tasks}
\label{sec:benchmarks}

We benchmark segmentation, depth estimation and completion, and 6D pose estimation. Within-dataset diagnostics use scene-disjoint splits and ten sampled viewpoints per trajectory; the full calibrated sequences remain available for video and multi-view methods. We also report results on ten held-out laboratory scenes with new layouts, distractors, backgrounds, and lighting, denoted by \textsc{TBiolab-HO}.

\subsection{Transparent-object segmentation}

In addition to the instance-matching SAM~3 protocol below, we compare SAM~3 \cite{carion2025sam3} and TransLab \cite{xie2020translab} using dataset-defined object masks. We report IoU, precision, recall, and F1; all four metrics are higher-is-better.

Table~\ref{tab:seg_crossdataset} shows that TransLab is strong on Trans10K but drops on TransBiolab and \textsc{TBiolab-HO}. SAM~3 also scores lower on TransBiolab than on ClearPose, indicating that the biomedical laboratory setting remains unsaturated.

\begin{table*}[t]
\caption{Mask-level segmentation on three datasets and the held-out scenes (higher is better).}
\label{tab:seg_crossdataset}
\centering
\small
\setlength{\tabcolsep}{8pt}
\begin{tabular}{@{}llcccc@{}}
\toprule
\textbf{Dataset / split} & \textbf{Method} & \textbf{IoU $\uparrow$} & \textbf{Precision $\uparrow$} & \textbf{Recall $\uparrow$} & \textbf{F1 $\uparrow$} \\
\midrule
TransBiolab-all & SAM~3~\cite{carion2025sam3} & 0.662 & 0.881 & 0.703 & 0.764 \\
TransBiolab-all & TransLab~\cite{xie2020translab} & 0.363 & 0.586 & 0.653 & 0.496 \\
ClearPose-all~\cite{chen2022clearpose} & SAM~3~\cite{carion2025sam3} & 0.699 & 0.885 & 0.744 & 0.789 \\
ClearPose-all~\cite{chen2022clearpose} & TransLab~\cite{xie2020translab} & 0.491 & 0.569 & 0.853 & 0.632 \\
Trans10K~\cite{xie2020translab} & SAM~3~\cite{carion2025sam3} & 0.704 & 0.821 & 0.827 & 0.824 \\
Trans10K~\cite{xie2020translab} & TransLab~\cite{xie2020translab} & 0.812 & 0.874 & 0.916 & 0.881 \\
\textsc{TBiolab-HO} & SAM~3~\cite{carion2025sam3} & 0.477 & 0.656 & 0.648 & 0.632 \\
\textsc{TBiolab-HO} & TransLab~\cite{xie2020translab} & 0.344 & 0.475 & 0.665 & 0.491 \\
\bottomrule
\end{tabular}
\end{table*}

\noindent\textbf{Instance-matching protocol.}
For each frame, we query SAM~3 with a text prompt for one object category and greedily match predicted masks to ground-truth instances by IoU. Unmatched ground-truth instances receive IoU 0. We report mIoU, match rate, and matched-instance mF1. On the evaluated subset, SAM~3 attains an overall mIoU of 0.685, a match rate of 84.8\%, and a matched-instance mF1 of 0.881.

\noindent\textbf{Object category.}
Table~\ref{tab:sam3obj} shows that dishes and plates are usually the easiest, with the Grex 1L bioreactor, 90 mm dish, 60 mm dish, and 24-well plate giving the strongest mIoU values. Tubes and larger flasks are harder, especially the 15 ml and 50 ml centrifuge tubes, whose thin cylindrical shapes and weak boundaries reduce both recall and overlap. Grouping the same results by geometric family gives the same ranking: plate-like objects perform best, bottle-like objects remain intermediate, and tube-like objects are the hardest.

\begin{table}[t]
\caption{SAM~3~\cite{carion2025sam3} object-category results.}
\label{tab:sam3obj}
\centering
\begin{minipage}{\narrowtabwidth}
\centering
\footnotesize
\setlength{\tabcolsep}{2.5pt}
\renewcommand{\arraystretch}{1.04}
\begin{tabularx}{\linewidth}{@{}>{\raggedright\arraybackslash}p{0.43\linewidth}>{\centering\arraybackslash}p{0.17\linewidth}>{\centering\arraybackslash}p{0.15\linewidth}>{\centering\arraybackslash}p{0.15\linewidth}@{}}
\toprule
\textbf{Name} & \textbf{Match rate} & \textbf{mIoU} & \textbf{mF1} \\
\midrule
Pipette & 89.9\% & 0.718 & 0.884 \\
Tube-rack & 90.0\% & 0.640 & 0.828 \\
60 mm dish & 95.5\% & 0.818 & 0.922 \\
6-well plate & 85.0\% & 0.766 & 0.946 \\
12-well plate & 87.8\% & 0.797 & 0.951 \\
24-well plate & 87.8\% & 0.802 & 0.953 \\
35 mm dish & 83.3\% & 0.720 & 0.900 \\
90 mm dish & 97.1\% & 0.845 & 0.925 \\
15 ml centrifuge tube & 71.7\% & 0.484 & 0.795 \\
50 ml centrifuge tube & 75.6\% & 0.495 & 0.780 \\
25 ml flask & 91.3\% & 0.735 & 0.883 \\
75 ml flask & 77.1\% & 0.574 & 0.829 \\
125 ml flask & 78.6\% & 0.587 & 0.799 \\
96-well plate & 81.4\% & 0.721 & 0.932 \\
Grex 1L bioreactor & 100.0\% & 0.914 & 0.954 \\
\midrule
Overall & 84.8\% & 0.685 & 0.881 \\
\bottomrule
\end{tabularx}
\end{minipage}
\end{table}

\begin{table}[t]
\caption{SAM~3~\cite{carion2025sam3} geometric-family results.}
\label{tab:sam3shape}
\centering
\begin{minipage}{\narrowtabwidth}
\centering
\footnotesize
\setlength{\tabcolsep}{2.5pt}
\renewcommand{\arraystretch}{1.04}
\begin{tabularx}{\linewidth}{@{}>{\raggedright\arraybackslash}p{0.22\linewidth}>{\raggedright\arraybackslash}p{0.42\linewidth}>{\centering\arraybackslash}p{0.14\linewidth}>{\centering\arraybackslash}p{0.16\linewidth}@{}}
\toprule
\textbf{Shape} & \textbf{Representative objects} & \textbf{mIoU} & \textbf{Match rate} \\
\midrule
Plate-like & 6/12/24/96-well plate and 35/60/90 mm dish & 0.789 & 89.2\% \\
Bottle-like & 25/75/125 ml flask, tube-rack, Grex 1L bioreactor & 0.700 & 86.3\% \\
Tube-like & 15/50 ml centrifuge tube and pipette & 0.536 & 76.1\% \\
\bottomrule
\end{tabularx}
\end{minipage}
\end{table}

\noindent\textbf{Scene clutter.}
We group frames by both the total number of visible objects and the number of categories. Performance is strongest in sparse settings and weaker in denser mixtures, but the decline is not strictly monotonic, suggesting interaction between clutter, geometry, and appearance.

\noindent\textbf{Viewpoint.}
Table~\ref{tab:sam3vp} shows a different pattern from pose estimation. From VP1 to VP8, mIoU remains relatively stable, while the main degradation appears only at the steepest views. At VP9 and VP10, match rate drops more than matched-instance mF1, indicating that missed detection dominates over matched-mask quality.

\begin{table}[t]
\caption{SAM~3~\cite{carion2025sam3} viewpoint results.}
\label{tab:sam3vp}
\centering
\begin{minipage}{\narrowtabwidth}
\centering
\footnotesize
\setlength{\tabcolsep}{1.8pt}
\renewcommand{\arraystretch}{1.04}
\begin{tabular*}{\linewidth}{@{\extracolsep{\fill}}lccccc@{}}
\toprule
\textbf{View} & \textbf{Elev.} & \textbf{Azim.} & \textbf{mIoU} & \textbf{Match} & \textbf{mF1} \\
\midrule
VP1 & -22.2 & -68.1 & 0.681 & 88.0\% & 0.856 \\
VP2 & -24.6 & -99.0 & 0.681 & 91.1\% & 0.837 \\
VP3 & -27.3 & -74.6 & 0.684 & 89.6\% & 0.847 \\
VP4 & -32.6 & -86.1 & 0.685 & 86.3\% & 0.874 \\
VP5 & -40.9 & -85.0 & 0.687 & 85.5\% & 0.880 \\
VP6 & -51.1 & -86.2 & 0.693 & 82.4\% & 0.905 \\
VP7 & -51.9 & -103.1 & 0.682 & 84.0\% & 0.886 \\
VP8 & -53.4 & -75.2 & 0.692 & 82.4\% & 0.905 \\
VP9 & -58.9 & -83.4 & 0.609 & 71.2\% & 0.919 \\
VP10 & -62.0 & -75.6 & 0.583 & 68.0\% & 0.915 \\
\bottomrule
\end{tabular*}
\end{minipage}
\end{table}

\subsection{Transparent-object depth estimation and completion}

We evaluate Depth Anything~3 (DA3) \cite{lin2025depthanything3} for metric-depth prediction and ClearGrasp \cite{sajjan2020cleargrasp} for transparent-object depth completion over dataset-defined object regions. DA3 uses metric depth without test-time scale or shift alignment. All reported metrics are lower-is-better.

\begin{table*}[t]
\caption{Object-region depth evaluation across datasets and on the held-out scenes. Lower values are better.}
\label{tab:depth_crossdataset}
\centering
\small
\setlength{\tabcolsep}{8pt}
\begin{tabular}{@{}lllcccc@{}}
\toprule
\textbf{Dataset} & \textbf{Split} & \textbf{Method} & \textbf{AbsRel $\downarrow$} & \textbf{RMSE (m) $\downarrow$} & \textbf{MAE (m) $\downarrow$} & \textbf{Raw missing $\downarrow$} \\
\midrule
TransBiolab & all & DA3~\cite{lin2025depthanything3} & 0.371 & 0.224 & 0.220 & 0.349 \\
TransBiolab & all & ClearGrasp~\cite{sajjan2020cleargrasp} & 0.393 & 0.473 & 0.349 & 0.313 \\
ClearPose~\cite{chen2022clearpose} & all & DA3~\cite{lin2025depthanything3} & 0.161 & 0.221 & 0.141 & 0.390 \\
ClearPose~\cite{chen2022clearpose} & all & ClearGrasp~\cite{sajjan2020cleargrasp} & 0.327 & 0.404 & 0.276 & 0.299 \\
TransBiolab & held-out & DA3~\cite{lin2025depthanything3} & 0.302 & 0.180 & 0.174 & 0.189 \\
TransBiolab & held-out & ClearGrasp~\cite{sajjan2020cleargrasp} & 0.179 & 0.132 & 0.086 & 0.178 \\
\bottomrule
\end{tabular}
\end{table*}

Both methods have higher object-region depth errors on TransBiolab-all than on ClearPose-all, suggesting that TransBiolab remains challenging for depth estimation and completion. ClearGrasp has lower held-out errors.

\begin{table*}[!t]
\caption{6D pose estimation across datasets and on the held-out laboratory scenes.}
\label{tab:pose_crossdataset}
\centering
\footnotesize
\setlength{\tabcolsep}{5pt}
\begin{tabular}{@{}lllcccc@{}}
\toprule
\textbf{Dataset / split} & \textbf{Method} & \textbf{Input} & \textbf{ADD-S AUC $\uparrow$} & \textbf{ADD AUC $\uparrow$} & \textbf{ADD-S mean $\downarrow$} & \textbf{Success $\uparrow$} \\
\midrule
TransBiolab-all & FoundationPose~\cite{wen2024foundationpose} & RGB-D, GT mask, CAD & 80.80 & 53.90 & 23.20 mm & 67.86\% \\
ClearPose-all~\cite{chen2022clearpose} & FoundationPose~\cite{wen2024foundationpose} & RGB-D, GT mask, CAD & 91.05 & 49.26 & 9.77 mm & 89.53\% \\
TransBiolab-all & MegaPose-6D~\cite{labbe2022megapose} & RGB, GT box, CAD & 31.76 & 31.49 & 67.76 mm & 34.14\% \\
ClearPose-all~\cite{chen2022clearpose} & MegaPose-6D~\cite{labbe2022megapose} & RGB, GT box, CAD & 39.84 & 12.40 & 64.29 mm & 25.00\% \\
\textsc{TBiolab-HO} & FoundationPose~\cite{wen2024foundationpose} & RGB-D, GT mask, CAD & 75.35 & 47.06 & 34.53 mm & 62.31\% \\
\textsc{TBiolab-HO} & MegaPose-6D~\cite{labbe2022megapose} & RGB, GT box, CAD & 55.50 & 25.17 & 96.46 mm & 26.59\% \\
\bottomrule
\end{tabular}
\end{table*}

\begin{table}[!t]
\caption{FoundationPose~\cite{wen2024foundationpose} by object category.}
\label{tab:objres}
\centering
\begin{minipage}{\narrowtabwidth}
\centering
\footnotesize
\setlength{\tabcolsep}{2.2pt}
\renewcommand{\arraystretch}{1.04}
\begin{tabularx}{\linewidth}{@{}>{\raggedright\arraybackslash}p{0.43\linewidth}>{\centering\arraybackslash}p{0.16\linewidth}>{\centering\arraybackslash}p{0.16\linewidth}>{\centering\arraybackslash}p{0.17\linewidth}@{}}
\toprule
\textbf{Name} & \textbf{ADD-S AUC} & \textbf{ADD AUC} & \textbf{ADD-S (mm)} \\
\midrule
Pipette & 93.0 & 88.2 & 10.8 \\
Tube-rack & 94.1 & 46.2 & 9.9 \\
60 mm dish & 90.1 & 56.9 & 9.9 \\
6-well plate & 91.7 & 49.2 & 8.6 \\
12-well plate & 91.9 & 44.2 & 8.1 \\
24-well plate & 93.4 & 52.8 & 6.6 \\
35 mm dish & 83.8 & 66.7 & 17.0 \\
90 mm dish & 90.6 & 37.6 & 11.0 \\
15 ml centrifuge tube & 75.4 & 64.4 & 24.8 \\
50 ml centrifuge tube & 81.0 & 58.7 & 19.7 \\
25 ml flask & 77.7 & 57.8 & 38.1 \\
75 ml flask & 54.2 & 35.0 & 53.2 \\
125 ml flask & 41.0 & 28.1 & 75.1 \\
96-well plate & 91.6 & 53.5 & 8.6 \\
Grex 1L bioreactor & 72.9 & 45.4 & 31.9 \\
\midrule
Overall & 80.8 & 53.9 & 23.2 \\
\bottomrule
\end{tabularx}
\end{minipage}
\end{table}

\begin{table}[!t]
\caption{FoundationPose~\cite{wen2024foundationpose} by geometric family.}
\label{tab:shape}
\centering
\begin{minipage}{\narrowtabwidth}
\centering
\footnotesize
\setlength{\tabcolsep}{2.2pt}
\renewcommand{\arraystretch}{1.04}
\begin{tabularx}{\linewidth}{@{}>{\raggedright\arraybackslash}p{0.22\linewidth}>{\raggedright\arraybackslash}p{0.44\linewidth}>{\centering\arraybackslash}p{0.14\linewidth}>{\centering\arraybackslash}p{0.14\linewidth}@{}}
\toprule
\textbf{Shape} & \textbf{Representative objects} & \textbf{ADD-S AUC} & \textbf{ADD-S (mm)} \\
\midrule
Plate-like & 6/12/24/96-well plate & 92.2 & 7.9 \\
Rack / elongated & Pipette, tube-rack & 93.6 & 10.4 \\
Dish-like & 35/60/90 mm dish & 90.4 & 10.4 \\
Tube-like & 15/50 ml centrifuge tube & 78.2 & 22.2 \\
Bottle-like & 25/75 ml flask & 65.9 & 45.6 \\
Large / irregular & 125 ml flask, Grex 1L bioreactor & 56.9 & 53.5 \\
\bottomrule
\end{tabularx}
\end{minipage}
\end{table}

\begin{table}[!t]
\caption{FoundationPose~\cite{wen2024foundationpose} results by viewpoint.}
\label{tab:vpres}
\centering
\begin{minipage}{\narrowtabwidth}
\centering
\footnotesize
\setlength{\tabcolsep}{1.8pt}
\renewcommand{\arraystretch}{1.04}
\begin{tabular*}{\linewidth}{@{\extracolsep{\fill}}lccccc@{}}
\toprule
\textbf{View} & \textbf{Elev.} & \textbf{Azim.} & \textbf{ADD-S AUC} & \textbf{ADD AUC} & \textbf{ADD-S} \\
\midrule
VP1 & -22.2 & -68.1 & 72.2 & 47.2 & 38.7 \\
VP2 & -24.6 & -99.0 & 72.2 & 43.2 & 48.4 \\
VP3 & -27.3 & -74.6 & 74.0 & 41.6 & 28.1 \\
VP4 & -32.6 & -86.1 & 73.9 & 48.7 & 30.1 \\
VP5 & -40.9 & -85.0 & 79.9 & 54.1 & 21.3 \\
VP6 & -51.1 & -86.2 & 88.0 & 60.3 & 12.2 \\
VP7 & -51.9 & -103.1 & 84.7 & 56.3 & 16.1 \\
VP8 & -53.4 & -75.2 & 87.1 & 60.3 & 13.3 \\
VP9 & -58.9 & -83.4 & 87.6 & 64.8 & 12.5 \\
VP10 & -62.0 & -75.6 & 87.8 & 61.5 & 12.5 \\
\bottomrule
\end{tabular*}
\end{minipage}
\end{table}

\subsection{6D object pose estimation}

FoundationPose is used as a generic baseline for 6D pose estimation \cite{wen2024foundationpose}. For each target instance, we load the RGB-D frame, visible mask, camera intrinsics, and CAD model, then run the standard register--refine pipeline with 15 refinement iterations. The stored dataset pose is converted to the object-in-camera convention used by the baseline. We report ADD, ADD-S, and AUC up to 0.1 m \cite{hinterstoisser2012linemod,xiang2018posecnn}. On the full dataset, FoundationPose reaches an overall ADD-S AUC of 80.8, an ADD AUC of 53.9, and a mean ADD-S of 23.2 mm.

For the added comparison, FoundationPose uses RGB-D, a ground-truth mask, and the CAD model, whereas MegaPose-6D \cite{labbe2022megapose} uses RGB, a ground-truth bounding box, and the CAD model without depth. Table~\ref{tab:pose_crossdataset} shows that FoundationPose drops from 91.05 ADD-S AUC on ClearPose-all to 80.80 on TransBiolab-all and 75.35 on \textsc{TBiolab-HO}. MegaPose-6D also remains limited without depth, suggesting that pose difficulty is not solely due to raw-depth artifacts. Because their inputs differ, the two methods do not form a controlled depth ablation.


\noindent\textbf{Object category.}
We first group results by object category. Table~\ref{tab:objres} lists all 15 objects, while Table~\ref{tab:shape} summarizes them by geometric family. Plates, racks, and dishes are generally easier, whereas tubes, flasks, and large irregular containers are harder. The ADD-S/ADD gap is also informative: tube racks, Petri dishes, and microplates keep high ADD-S AUC but much lower ADD AUC, indicating uncertainty in rotational identity under symmetry. By contrast, the pipette has a smaller gap because its asymmetric geometry offers stronger directional cues. These patterns show that \datasetname{} exposes difficulty tied to transparency, geometry, and symmetry rather than clutter alone.

\begin{figure}[t]
    \centering
    \includegraphics[width=\linewidth]{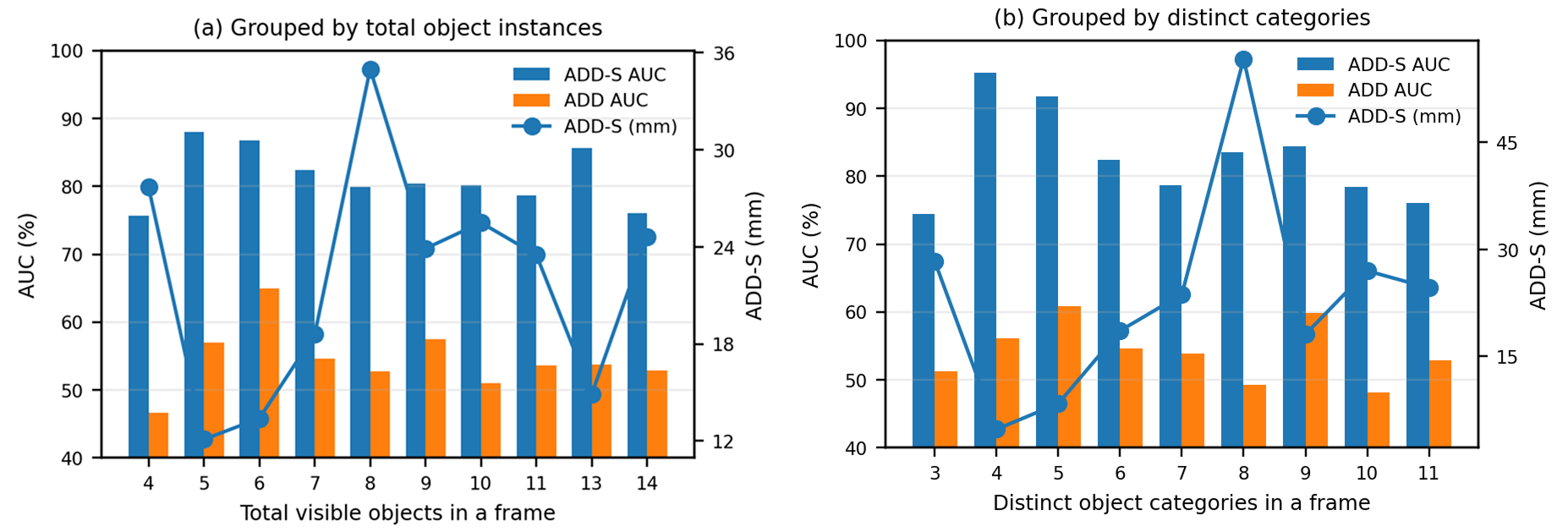}
    \caption{Pose results by scene complexity.}
    \Description{Two plots summarizing FoundationPose performance against scene complexity, using total object count and distinct category count as grouping axes.}
    \label{fig:clutter}
\end{figure}

\noindent\textbf{Scene clutter.}
We next group results by both the total number of visible objects and the number of distinct categories in each frame. Figure~\ref{fig:clutter} shows that the trend is not strictly monotonic: some moderate-clutter bins outperform sparser ones because geometry and symmetry still dominate difficulty. This indicates that clutter interacts with partial visibility and object shape rather than behaving as a single scalar factor.

\noindent\textbf{Viewpoint.}
Table~\ref{tab:vpres} shows a clear viewpoint effect. Shallow views are the hardest: VP1 starts at ADD-S AUC 72.2 with mean ADD-S 38.7 mm, whereas VP10 reaches ADD-S AUC 87.8 with mean ADD-S 12.5 mm. Mid-to-steep views are more stable because they reveal larger top surfaces and clearer contours, suggesting that view planning and multi-view fusion remain promising.


\subsection{System-level real-robot evaluation}

\begin{figure}[t]
    \centering
    \includegraphics[width=\linewidth]{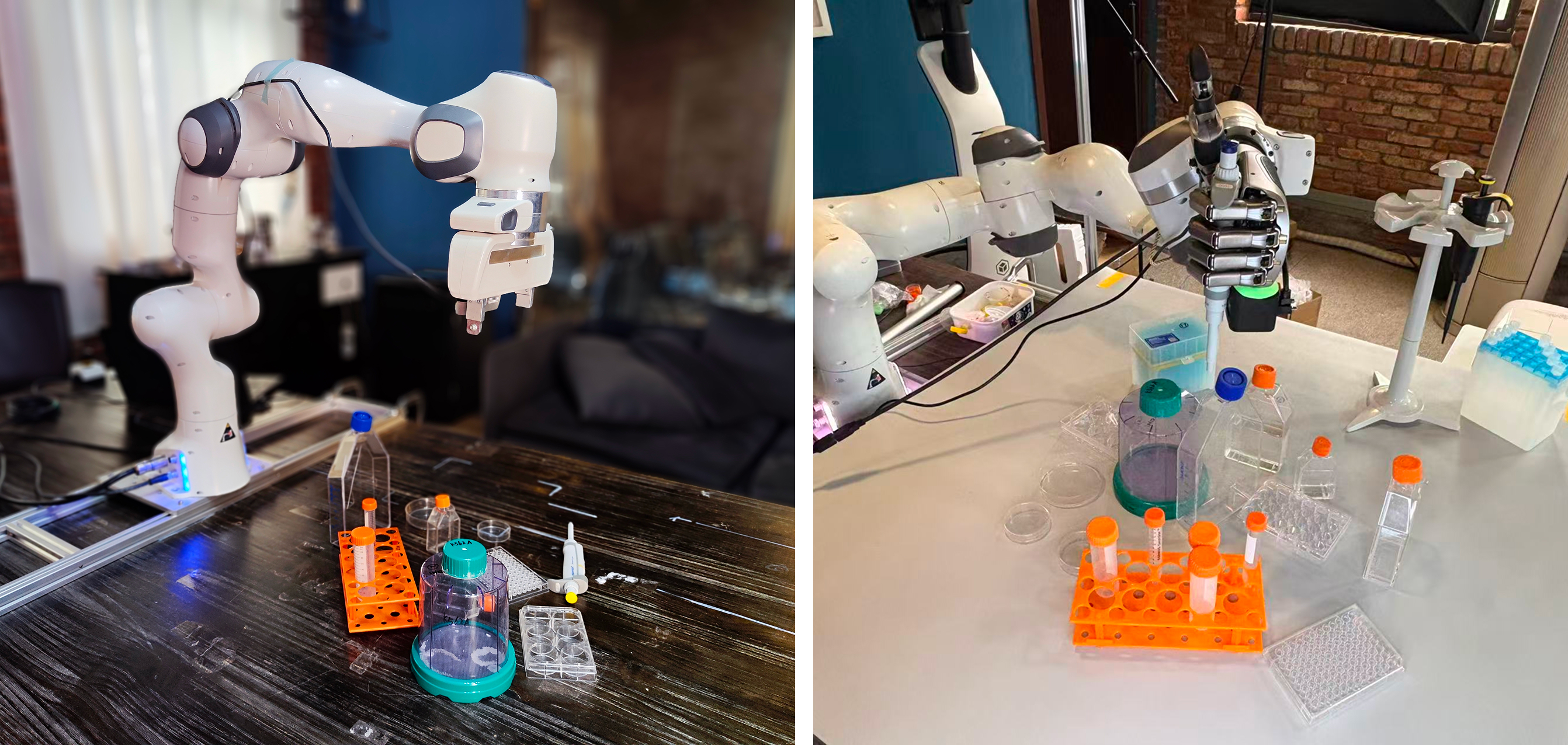}
    \caption{Robot manipulation platform.}
    \Description{Left: a robot arm with an RGB-D camera above a tabletop scene containing several transparent laboratory objects. Right: Dexterous-hand manipulation setting driven by the released segmentation and pose annotations.}
    \label{fig:robotsetup}
\end{figure}

\datasetname{} is also intended to support end-to-end manipulation experiments on a real robot platform. As shown in Figure~\ref{fig:robotsetup}, the system first obtains an object mask from segmentation, then estimates a target 6D pose with FoundationPose, transforms this pose to the robot base through hand-eye calibration, and finally plans grasp and placement motions through ROS/MoveIt. The same 6D pose output can also initialize dexterous-hand grasp synthesis.

We execute real robot grasping trials in cluttered tabletop scenes using two end-effectors. With a Franka parallel-jaw gripper, the system succeeds 98 times out of 150 trials, for a success rate of 65.3\%. Replacing the end-effector with a LinkerHand 10-DoF dexterous hand yields 85 successful grasps out of 150 trials, for a success rate of 56.67\%. This is a system-level manipulation evaluation rather than a proxy for 6D pose accuracy. Compared with parallel-jaw grasping, LinkerHand additionally requires grasp-type selection, high-dimensional joint mapping, and contact-rich closure; its lower success may therefore reflect grasp synthesis and control complexity in addition to perception error.

\section{Conclusion}

We presented \datasetname{}, a real-world multi-view dataset of cluttered transparent biomedical objects. \datasetname{} contains 161{,}315 frames from 98 scenes and 1.03M instance annotations over 15 laboratory object types, including 6D poses, full and visible masks, depth, and per-frame camera calibration. Across segmentation, depth estimation and completion, and 6D pose estimation, the expanded results show that current methods still exhibit substantial gaps across domains, object geometry, symmetry, clutter, and viewpoint. The robot experiments provide a system-level manipulation evaluation rather than a proxy for pose accuracy, because the measured success also reflects grasp synthesis and control. We therefore view \datasetname{} as a dataset for studying unresolved boundary cases rather than a solved benchmark. Upon publication, we will release the full RGB-D sequences, 15 OBJ CAD models, 6D poses, masks, depth, calibration files, and evaluation scripts. The dataset does not contain faces, personal identities, or patient data.

\bibliographystyle{ACM-Reference-Format}
\bibliography{software}

@article{abolhasani2023rise,
  author    = {Abolhasani, Milad and Kumacheva, Eugenia},
  title     = {The rise of self-driving labs in chemical and materials sciences},
  journal   = {Nature Synthesis},
  year      = {2023},
  volume    = {2},
  pages     = {483--492}
}

@article{tom2024selfdriving,
  author    = {Tom, Gary and Schmid, Stefan P. and Baird, Sterling G. and Wang, Zhi and Choi, Livia and Kumar, Naveen and Aspuru-Guzik, Alan},
  title     = {Self-driving laboratories for chemistry and materials science},
  journal   = {Chemical Reviews},
  year      = {2024},
  volume    = {124},
  number    = {16},
  pages     = {9633--9732}
}

@article{volk2024performance,
  author    = {Volk, Amanda A. and Abolhasani, Milad},
  title     = {Performance metrics to unleash the power of self-driving labs in chemistry and materials science},
  journal   = {Nature Communications},
  year      = {2024},
  volume    = {15},
  pages     = {1378}
}

@article{fushimi2025autonomous,
  author    = {Fushimi, Keiji and Nakai, Yusuke and Nishi, Akiko and Suzuki, Ryo and Ikegami, Masahiro and Nimura, Risa and Tomono, Taichi and Hidese, Ryota and Yasueda, Hisashi and Tagawa, Yusuke and others},
  title     = {Development of the autonomous lab system to support biotechnology research},
  journal   = {Scientific Reports},
  year      = {2025},
  volume    = {15},
  pages     = {6648}
}

@article{melocchi2024roboticcluster,
  author    = {Melocchi, Alice and Schmittlein, Brigitte and Jones, Alexis L. and Ainane, Yasmine and Rizvi, Ali and Chan, Darius and Dickey, Elaine and others},
  title     = {Development of a robotic cluster for automated and scalable cell therapy manufacturing},
  journal   = {Cytotherapy},
  year      = {2024},
  volume    = {26},
  number    = {9},
  pages     = {1095--1104}
}

@article{ochs2021stemcelldiscovery,
  author    = {Ochs, Julia and Golkaram, Mahyar and Joussen, Silja and Wittemeier, Alexander and Frank, Max and M{\"o}llmann, Christina and others},
  title     = {Fully automated cultivation of adipose-derived stem cells in the {StemCellDiscovery}---a robotic laboratory for small-scale, high-throughput cell production including deep learning-based confluence estimation},
  journal   = {Stem Cell Reviews and Reports},
  year      = {2021},
  volume    = {17},
  pages     = {1651--1668}
}

@article{moutsatsou2019automation,
  author    = {Moutsatsou, Polyxeni and Ochs, Julia and Schmitt, Robert H. and Hewitt, Christopher J. and Hanga, Mariana P.},
  title     = {Automation in cell and gene therapy manufacturing: from past to future},
  journal   = {Biotechnology Letters},
  year      = {2019},
  volume    = {41},
  number    = {11--12},
  pages     = {1245--1253}
}

@article{doulgkeroglou2020automation,
  author    = {Doulgkeroglou, Maria-Nefeli and Di Nubila, Andrea and Niessing, Bianca and Kearns, Verena and Schmitt, Robert H. and Martin, Ingo and de Wildt, Bram and Spee, Bjorn and others},
  title     = {Automation, monitoring, and standardization of cell product manufacturing},
  journal   = {Frontiers in Bioengineering and Biotechnology},
  year      = {2020},
  volume    = {8},
  pages     = {811}
}

@inproceedings{zhang2023manualpipette,
  author    = {Zhang, Yuhang and Ishikawa, Ryohei and Tanaka, Tsuyoshi and Hashimoto, Koichi and Oikawa, Hiroyuki},
  title     = {Integrating a manual pipette into a collaborative robot manipulator for flexible liquid dispensing},
  booktitle = {IEEE/SICE International Symposium on System Integration (SII)},
  year      = {2023},
  pages     = {1--6}
}

@inproceedings{hinterstoisser2012linemod,
  author    = {Hinterstoisser, Stefan and Lepetit, Vincent and Ilic, Slobodan and Holzer, Stefan and Bradski, Gary and Konolige, Kurt and Navab, Nassir},
  title     = {Model based training, detection and pose estimation of texture-less 3D objects in heavily cluttered scenes},
  booktitle = {Asian Conference on Computer Vision (ACCV)},
  year      = {2012},
  pages     = {548--562}
}

@inproceedings{hodan2018bop,
  author    = {Hodan, Tomas and Michel, Frank and Brachmann, Eric and Kehl, Wadim and Glent Buch, Anders and Kraft, Dirk and Doumanoglou, Anastasios and others},
  title     = {{BOP}: Benchmark for 6D object pose estimation},
  booktitle = {European Conference on Computer Vision (ECCV)},
  year      = {2018},
  pages     = {19--35}
}

@article{hodan2017tless,
  author    = {Hodan, Tomas and Haluza, Pavel and Obdrzalek, Stepan and Matas, Jiri and Lourakis, Manolis and Zabulis, Xenophon},
  title     = {{T-LESS}: An RGB-D dataset for 6D pose estimation of texture-less objects},
  journal   = {IEEE Winter Conference on Applications of Computer Vision (WACV)},
  year      = {2017},
  pages     = {880--888}
}

@inproceedings{drost2017itodd,
  author    = {Drost, Bertram and Ulrich, Markus and Nitschke, Christian and others},
  title     = {Introducing {MVTec ITODD}---a dataset for 3D object recognition in industry},
  booktitle = {IEEE International Conference on Computer Vision Workshops (ICCVW)},
  year      = {2017}
}

@inproceedings{rennie2016ruapc,
  author    = {Rennie, Colin and Shome, Rohan and Bekris, Kostas and De Souza, Alberto F.},
  title     = {A dataset for improved RGBD-based object detection and pose estimation for warehouse pick-and-place},
  booktitle = {IEEE International Conference on Robotics and Automation Workshops (ICRAW)},
  year      = {2016}
}

@inproceedings{doumanoglou2016recovering,
  author    = {Doumanoglou, Anastasios and Kouskouridas, Rigas and Kim, Tae-Kyun and Malassiotis, Sotiris and Komodakis, Nikos},
  title     = {Recovering 6D object pose and predicting next-best-view in the crowd},
  booktitle = {IEEE Conference on Computer Vision and Pattern Recognition (CVPR)},
  year      = {2016},
  pages     = {3583--3592}
}

@inproceedings{calli2015ycb,
  author    = {Calli, Berk and Singh, Arjun and Walsman, Aaron and Srinivasa, Siddhartha and Abbeel, Pieter and Dollar, Aaron M.},
  title     = {Benchmarking in manipulation research: The {YCB} object and model set and benchmarking protocols},
  booktitle = {IEEE International Conference on Advanced Robotics (ICAR)},
  year      = {2015},
  pages     = {510--517}
}

@inproceedings{xiang2018posecnn,
  author    = {Xiang, Yu and Schmidt, Tanner and Narayanan, Venkatraman and Fox, Dieter},
  title     = {{PoseCNN}: A convolutional neural network for 6D object pose estimation in cluttered scenes},
  booktitle = {Robotics: Science and Systems (RSS)},
  year      = {2018}
}

@inproceedings{fang2020graspnet,
  author    = {Fang, Hao-Shu and Wang, Chenxi and Gou, Minghao and Lu, Cewu},
  title     = {{GraspNet-1Billion}: A large-scale benchmark for general object grasping},
  booktitle = {IEEE/CVF Conference on Computer Vision and Pattern Recognition (CVPR)},
  year      = {2020},
  pages     = {11444--11453}
}

@inproceedings{tyree2022hope,
  author    = {Tyree, Stephen and Tremblay, Jonathan and To, Tommy and Cheng, Jia and Mosier, Terry and Smith, Jeffrey and Birchfield, Stan and Hager, Gregory D.},
  title     = {6-DoF pose estimation of household objects for robotic manipulation: an accessible dataset and benchmark},
  booktitle = {Conference on Robot Learning (CoRL)},
  year      = {2022}
}

@inproceedings{yang2021robi,
  author    = {Yang, Jun and Gao, Yizhou and Li, Dong and Waslander, Steven L.},
  title     = {{ROBI}: A multi-view dataset for reflective objects in robotic bin-picking},
  booktitle = {IEEE/RSJ International Conference on Intelligent Robots and Systems (IROS)},
  year      = {2021}
}

@inproceedings{liu2020keypose,
  author    = {Liu, Xingyu and Jonschkowski, Rico and Angelova, Anelia and Konolige, Kurt},
  title     = {KeyPose: Multi-view 3D labeling and keypoint estimation for transparent objects},
  booktitle = {IEEE/CVF Conference on Computer Vision and Pattern Recognition (CVPR)},
  year      = {2020},
  pages     = {11602--11610}
}

@inproceedings{sajjan2020cleargrasp,
  author    = {Sajjan, Nikhil and Moore, Matthew and Pan, Mikey and Nagaraja, Gaurav and Lee, Jia and Zeng, Andy and Song, Shuran and Funkhouser, Thomas},
  title     = {{ClearGrasp}: 3D shape estimation of transparent objects for manipulation},
  booktitle = {IEEE International Conference on Robotics and Automation (ICRA)},
  year      = {2020},
  pages     = {3634--3642}
}

@inproceedings{xu2022seeingglass,
  author    = {Xu, Haoping and Wang, Yi Ru and Eppel, Sagi and Aspuru-Guzik, Alan and Shkurti, Florian and Garg, Animesh},
  title     = {Seeing glass: Joint point-cloud and depth completion for transparent objects},
  booktitle = {Conference on Robot Learning (CoRL)},
  year      = {2022},
  series    = {Proceedings of Machine Learning Research},
  volume    = {164},
  pages     = {827--838}
}

@inproceedings{liu2021stereobj,
  author    = {Liu, Xingyu and Iwase, Shun and Kitani, Kris M.},
  title     = {{StereOBJ}-1M: Large-scale stereo image dataset for 6D object pose estimation},
  booktitle = {IEEE/CVF International Conference on Computer Vision Workshops (ICCVW)},
  year      = {2021},
  pages     = {1294--1303}
}

@inproceedings{chen2022clearpose,
  author    = {Chen, Xiaotong and Zhang, Huijie and Yu, Zeren and Opipari, Anthony and Jenkins, Odest Chadwicke},
  title     = {ClearPose: Large-scale transparent object dataset and benchmark},
  booktitle = {European Conference on Computer Vision (ECCV)},
  year      = {2022},
  pages     = {381--398}
}

@article{fang2022transcg,
  author    = {Fang, Hongjie and Fang, Hao-Shu and Xu, Sheng and Lu, Cewu},
  title     = {{TransCG}: A large-scale real-world dataset for transparent object depth completion and a grasping baseline},
  journal   = {IEEE Robotics and Automation Letters},
  year      = {2022},
  volume    = {7},
  number    = {3},
  pages     = {7383--7390}
}

@inproceedings{xie2020translab,
  author    = {Xie, Enze and Wang, Wenjia and Wang, Wenhai and Ding, Mingyu and Shen, Chunhua and Luo, Ping},
  title     = {Segmenting Transparent Objects in the Wild},
  booktitle = {European Conference on Computer Vision (ECCV)},
  year      = {2020},
  pages     = {696--711}
}

@inproceedings{xie2021trans2seg,
  author    = {Xie, Enze and Wang, Wenjia and Wang, Wenhai and Sun, Peize and Xu, Hang and Liang, Ding and Luo, Ping},
  title     = {Segmenting transparent object in the wild with transformer},
  booktitle = {International Joint Conference on Artificial Intelligence (IJCAI)},
  year      = {2021},
  pages     = {1003--1009}
}

@article{lukezic2024transparenttracking,
  author    = {Lukezic, Alan and Trojer, Ziga and Matas, Jiri and others},
  title     = {A new dataset and a distractor-aware architecture for transparent object tracking},
  journal   = {International Journal of Computer Vision},
  year      = {2024},
  volume    = {132},
  pages     = {2729--2742}
}

@inproceedings{wen2024foundationpose,
  author    = {Wen, Bowen and Yang, Wei and Kautz, Jan and Birchfield, Stan},
  title     = {FoundationPose: Unified 6D pose estimation and tracking of novel objects},
  booktitle = {IEEE/CVF Conference on Computer Vision and Pattern Recognition (CVPR)},
  year      = {2024},
  pages     = {17868--17879}
}

@inproceedings{labbe2022megapose,
  author    = {Labb{\'e}, Yann and Manuelli, Lucas and Mousavian, Arsalan and Tyree, Stephen and Birchfield, Stan and Tremblay, Jonathan and Carpentier, Justin and Aubry, Mathieu and Fox, Dieter and Sivic, Josef},
  title     = {MegaPose: 6D Pose Estimation of Novel Objects via Render \& Compare},
  booktitle = {Conference on Robot Learning (CoRL)},
  year      = {2022}
}

@article{ravi2024sam2,
  author    = {Ravi, Nikhila and Gabeur, Valentin and Hu, Yuan-Ting and Hu, Ronghang and Ryali, Chaitanya and Gustafson, Laura and others},
  title     = {{SAM} 2: Segment anything in images and videos},
  journal   = {arXiv preprint arXiv:2408.00714},
  year      = {2024}
}

@article{carion2025sam3,
  author    = {Carion, Nicolas and Gustafson, Laura and Hu, Yuan-Ting and Debnath, Shoubhik and Hu, Ronghang and Suris, Didac and Ryali, Chaitanya and Alwala, Kalyan Vasudev and others},
  title     = {{SAM} 3: Segment anything with concepts},
  journal   = {arXiv preprint arXiv:2511.16719},
  year      = {2025}
}

@article{yang2024depthanythingv2,
  author    = {Yang, Lihe and Kang, Bingyi and Huang, Zilong and Xu, Xin and Feng, Jiashi and Zhao, Hengshuang},
  title     = {Depth Anything V2},
  journal   = {arXiv preprint arXiv:2406.09414},
  year      = {2024}
}

@article{lin2025depthanything3,
  author    = {Lin, Haotong and Chen, Sili and Liew, Junhao and Chen, Donny Y. and Li, Zhenyu and Shi, Guang and Feng, Jiashi and Kang, Bingyi},
  title     = {Depth Anything 3: Recovering the visual space from any views},
  journal   = {arXiv preprint arXiv:2511.10647},
  year      = {2025}
}

@inproceedings{chen2022progresslabeller,
  author    = {Chen, Xiaotong and Zhang, Huijie and Yu, Zeren and Lewis, Stanley and Jenkins, Odest Chadwicke},
  title     = {ProgressLabeller: Visual data stream annotation for training object-centric 3D perception},
  booktitle = {IEEE/RSJ International Conference on Intelligent Robots and Systems (IROS)},
  year      = {2022},
  pages     = {13066--13073}
}

@article{campos2021orbslam3,
  author    = {Campos, Carlos and Elvira, Richard and Rodr{\'i}guez, Juan J. G{\'o}mez and Montiel, Jos{\'e} M. M. and Tard{\'o}s, Juan D.},
  title     = {{ORB-SLAM3}: An accurate open-source library for visual, visual-inertial, and multimap {SLAM}},
  journal   = {IEEE Transactions on Robotics},
  year      = {2021},
  volume    = {37},
  number    = {6},
  pages     = {1874--1890}
}

@inproceedings{ma2026phys,
  title        = {Phys-liquid: a physics-informed dataset for estimating 3d geometry and volume of transparent deformable liquids},
  author       = {Ma, Ke and Fang, Yizhou and Weibel, Jean-Baptiste and Tan, Shuai and Wang, Xinggang and Xiao, Yang and Fang, Yi and Xia, Tian},
  booktitle    = {Proceedings of the AAAI Conference on Artificial Intelligence},
  volume       = {40},
  number       = {10},
  pages        = {7782--7790},
  year         = {2026}
}

@inproceedings{tan2026synmotion,
  title        = {Synmotion: Semantic-visual adaptation for motion customized video generation},
  author       = {Tan, Shuai and Gong, Biao and Wei, Yujie and Zhang, Shiwei and Liu, Zhuoxin and Ma, Ke and Wang, Yan and Zheng, Kecheng and Zhu, Xing and Shen, Yujun and others},
  booktitle    = {Proceedings of the IEEE/CVF Conference on Computer Vision and Pattern Recognition},
  pages        = {30477--30489},
  year         = {2026}
}

@inproceedings{ma2019design,
  title        = {Design pattern as a practical tool for designing adaptive interactions connecting human and social robots},
  author       = {Ma, Ke and Cao, Jing},
  booktitle    = {International Conference on Intelligent Human Systems Integration},
  pages        = {613--617},
  year         = {2019},
  organization = {Springer}
}

@article{tan2026codance,
  title        = {{CoDance}: An Unbind-Rebind Paradigm for Robust Multi-Subject Animation},
  author       = {Tan, Shuai and Gong, Biao and Ma, Ke and Feng, Yutong and Zhang, Qiyuan and Wang, Yan and Shen, Yujun and Zhao, Hengshuang},
  journal      = {arXiv preprint arXiv:2601.11096},
  year         = {2026}
}

@inproceedings{ma2026wpis,
  title        = {{WPIS}: From In-the-Wild Web Images to Physics-Aware 3D Scene Graphs for Physical Reasoning},
  author       = {Ma, Ke and Fu, Cong and Wang, Jianing and Wang, Yifei and Li, Wenyuan and Wang, Xinggang and Wang, Meng and Xia, Tian},
  booktitle    = {Proceedings of the ACM Web Conference 2026},
  pages        = {1410--1421},
  year         = {2026}
}

@inproceedings{shi2017digital,
  title        = {Digital touchpoints in campus slow traffic service system},
  author       = {Shi, Jintian and Ma, Ke},
  booktitle    = {International Conference on Applied Human Factors and Ergonomics},
  pages        = {349--361},
  year         = {2017},
  organization = {Springer}
}

@inproceedings{tan2024edtalk,
  title        = {{EDTalk}: Efficient disentanglement for emotional talking head synthesis},
  author       = {Tan, Shuai and Ji, Bin and Bi, Mengxiao and Pan, Ye},
  booktitle    = {European Conference on Computer Vision},
  pages        = {398--416},
  year         = {2024},
  organization = {Springer}
}

@inproceedings{tan2025fixtalk,
  title        = {{FixTalk}: Taming identity leakage for high-quality talking head generation in extreme cases},
  author       = {Tan, Shuai and Gong, Bill and Ji, Bin and Pan, Ye},
  booktitle    = {Proceedings of the IEEE/CVF International Conference on Computer Vision},
  pages        = {24--36},
  year         = {2025}
}

\end{document}